\documentclass[conference]{IEEEtran}
\IEEEoverridecommandlockouts
\usepackage{cite}
\usepackage{amsmath,amssymb,amsfonts}
\usepackage{graphicx}
\usepackage{textcomp}
\usepackage{xcolor}
\usepackage{multirow}
\usepackage{float}

\def\BibTeX{{\rm B\kern-.05em{\sc i\kern-.025em b}\kern-.08em
    T\kern-.1667em\lower.7ex\hbox{E}\kern-.125emX}}

\newcommand{\authcell}[5]{%
  \makebox[2.05in][c]{\shortstack{#1\\[1pt] \textit{#2}\\ \textit{#3}\\ #4\\ #5}}}
\begin{document}

\title{Comprehensive Benchmarking of Deep Learning Architectures for Lung Cancer Histopathology}

\author{%
\small
\begin{tabular}{c@{\hspace{0.8em}}c@{\hspace{0.8em}}c}
\authcell{Hadi Hasan}{Electrical and Computer Engineering}{American University of Beirut}{Beirut, Lebanon}{hsh24@mail.aub.edu} &
\authcell{Safaa Salman}{Electrical and Computer Engineering}{American University of Beirut}{Beirut, Lebanon}{sns44@mail.aub.edu} &
\authcell{Lama Sleem}{Computational Science}{American University of Beirut}{Beirut, Lebanon}{lks09@mail.aub.edu} \\[1.4em]
\multicolumn{3}{c}{%
\authcell{Ralph Mouawad}{Industrial Engineering}{American University of Beirut}{Beirut, Lebanon}{rmm85@mail.aub.edu}\hspace{0.8em}%
\authcell{Ali Chehab}{Electrical and Computer Engineering}{American University of Beirut}{Beirut, Lebanon}{chehab@aub.edu.lb}} \\
\end{tabular}}

\maketitle

\begin{abstract}
Lung cancer remains the leading cause of cancer-related mortality worldwide, while histopathological diagnosis is often affected by inter-observer variability and the substantial workload associated with manual slide examination. Although deep learning has shown considerable potential in computational pathology, comprehensive benchmarks that integrate tissue classification and region segmentation within a unified analytical framework remain limited. This study presents a two-stage deep learning framework for multi-class tissue classification and pixel-level histopathological region segmentation, accompanied by a systematic comparison of state-of-the-art architectures at each stage. For tissue classification, six models, a custom convolutional neural network, VGG16, DenseNet, MobileNetV3, a custom Vision Transformer, and YOLO11, are evaluated on a combined dataset of 39,000 images derived from LC25000 and LungHist700. The models distinguish between adenocarcinoma, squamous cell carcinoma, and normal lung tissue. YOLO11 achieves the best classification performance, with an accuracy of 98.38\%, a five-fold cross-validation accuracy of $98.21 \pm 0.35\%$, and a macro F1-score of 0.98. For region segmentation, U-Net, ResNet-encoder U-Net, DeepLabV3+, and YOLO11-seg are evaluated using the GlaS gland segmentation benchmark. DeepLabV3+ obtains the highest Intersection over Union of 0.80 and a Dice score of 0.89, while YOLO11-seg achieves a comparable Intersection over Union of 0.79 using approximately $14\times$ fewer parameters. The best-performing classification and segmentation models are subsequently integrated into an end-to-end framework, providing an accurate, computationally efficient, and reproducible baseline for automated histopathological image analysis.
\end{abstract}

\begin{IEEEkeywords}
Computational pathology, lung cancer, deep learning, histopathology, image classification, image segmentation, YOLO11, DeepLabV3+, transfer learning, benchmark.
\end{IEEEkeywords}

\section{Introduction}
\label{sec:introduction}
Histopathological analysis is the gold standard for definitive cancer diagnosis, yet its accuracy depends heavily on pathologist expertise and is subject to significant inter-observer variability, arising from interpretive subjectivity and the cognitive burden of examining many slides per session~\cite{gurcan2009histopathological, basu2024survey}. The challenge is acute for lung cancer, the deadliest malignancy worldwide~\cite{barta2019global}, where images from biopsy specimens require both tissue classification (ACA, SCC, or normal) and spatial delineation of pathological regions, and where manual analysis is labor intensive and susceptible to variability arising from staining protocols, tissue preparation, and intra-tumoral heterogeneity~\cite{moscalu2023histopathological}. Automated systems that reliably perform both tasks within an integrated framework are therefore of considerable clinical value; we have previously found such a cascade practical for opportunistic screening on routine CT~\cite{hasan2026end}.

Deep learning has shown substantial promise for automating histopathological analysis, offering consistent and reproducible predictions at scale~\cite{basu2024survey}. Three gaps persist, however. Most studies address classification and segmentation as isolated tasks, with few benchmarks integrating both into a unified pipeline; multi-architecture comparisons spanning convolutional, transformer-based, and detection-based paradigms under a single controlled protocol remain rare, complicating principled model selection; and versatile recent architectures such as YOLO11, originally designed for object detection, have not been thoroughly evaluated for histopathology classification or segmentation. Accuracy and cost are also necessary but not sufficient for clinical use, which additionally demands robustness under distribution shift and auditable decisions~\cite{hasan2026defending, hasan2026toward}.

This study addresses these gaps through a systematic benchmarking and integration study, and we state the nature of that contribution explicitly: every model evaluated here is an established architecture, and the novelty lies not in a new network design but in the controlled comparative evidence produced and in the protocol that converts that evidence into a working two-stage pipeline. Benchmarking in computational pathology has largely compared models within a single task and a single architectural family---most commonly convolutional classifiers on LC25000---leaving unresolved how paradigms compare when classification and segmentation must be selected jointly under a shared computational budget. This work instead holds preprocessing, augmentation, optimizer, schedule, data partitioning, cross-validation, and hardware fixed across all ten configurations, so that observed differences are attributable to architecture rather than to training-recipe variation; reports accuracy jointly with parameter count, GFLOPs, and measured latency, posing model selection as an accuracy--efficiency trade-off rather than a leaderboard ranking; and characterizes the detection-family YOLO11 and YOLO11-seg on tissue sub-typing and gland segmentation, tasks for which this family has not been systematically evaluated. The classification stage uses a merged dataset of 39,000 images from LC25000~\cite{borkowski2019lung} and LungHist700~\cite{Diosdado2024LungHist700}; the segmentation stage uses GlaS~\cite{sirinukunwattana2017gland}, as lung-specific pixel-level annotations remain extremely scarce. The principal contributions are:
\begin{enumerate}
    \item \textbf{A single-protocol benchmark spanning three architectural paradigms.} Six classifiers and four segmentation models are trained and evaluated under identical data, augmentation, optimization, and hardware conditions, with per-class metrics, stratified 5-fold cross-validation, and cost profiling reported for every configuration (Table~\ref{tab:hyperparams}).
    \item \textbf{Transferable empirical findings rather than a single accuracy figure.} The benchmark establishes that detection-pretrained backbones transfer effectively to histopathology sub-typing (YOLO11, 98.38\%); that encoder pretraining rather than decoder sophistication dominates segmentation quality on small annotated sets ($+0.27$ IoU from U-Net to ResNet-U-Net, against $+0.02$ from ResNet-U-Net to DeepLabV3+); and that returns diminish beyond a ResNet-50 encoder, identifying annotation volume rather than model capacity as the binding constraint.
    \item \textbf{An accuracy--efficiency characterization for deployment-oriented model selection.} Reporting parameters, GFLOPs, and latency alongside accuracy exposes trade-offs invisible to accuracy-only benchmarks, most notably that YOLO11-seg reaches within 0.01 IoU of DeepLabV3+ using approximately $14\times$ fewer parameters.
    \item \textbf{An integrated and reproducible two-stage pipeline.} The best classifier (YOLO11) is cascaded with the best segmentation model (DeepLabV3+), supported by ablations isolating dataset merging, augmentation, and encoder backbone, and by a leakage-controlled parent-image splitting procedure.
\end{enumerate}

\section{Clinical Background}
\label{sec:background}
Histopathology is the microscopic examination of tissue obtained by biopsy or resection. The specimen is fixed, embedded in paraffin, sectioned at a few micrometers, and stained---most commonly with hematoxylin and eosin (H\&E), which renders nuclei blue-purple and cytoplasm and stroma pink---before examination under a microscope or, increasingly, as a digitized whole-slide image~\cite{gurcan2009histopathological}. Lung cancer is divided into small cell lung carcinoma and non-small cell lung carcinoma (NSCLC), the latter accounting for the large majority of cases~\cite{barta2019global}. The two dominant NSCLC subtypes are adenocarcinoma (ACA), which arises from glandular epithelium and typically exhibits gland formation or lepidic growth, and squamous cell carcinoma (SCC), which arises from bronchial squamous epithelium and is recognized by keratinization and intercellular bridges. The distinction is consequential rather than taxonomic: ACA and SCC diverge in eligibility for targeted therapy, immunotherapy, and specific chemotherapeutic agents, so subtype assignment directly conditions treatment~\cite{nooreldeen2021current}. In routine practice the pathologist screens the slide at low magnification to locate suspicious regions, then examines them at high magnification to assign subtype and grade. The two operations at the core of this workflow---deciding which tissue type is present, and delineating where the abnormal tissue lies---map directly onto the classification and segmentation stages of the pipeline proposed here.

\section{Related Work}
\textbf{Histopathology image classification.}
Early deep-learning approaches employed task-specific convolutional neural networks~\cite{tajbakhsh2016cnn}. Abbas et al.~\cite{abbas2020histopathological} compared six pretrained CNNs on the three-class lung subset of LC25000 and reported F1-scores of 97.3--99.9\%, while Humayun et al.~\cite{humayun2022transfer} reached 98.83\% with VGG16 on the same source through transfer learning~\cite{pan2010transfer}; fusing CNN and handcrafted features has yielded further gains on morphologically complex tissue such as malignant lymphoma~\cite{hamdi2023hybrid}. Work published in 2025 has continued this trajectory while beginning to saturate the standard benchmark. Ochoa-Ornelas et al.~\cite{ochoa2025robust} reported 99.39\% with EfficientNetB3 on LC25000, a figure that leaves little headroom and illustrates why single-source evaluation on this dataset is increasingly uninformative for model selection. Nahmatwlla and Ali~\cite{nahmatwlla2025hybrid} augmented a ConvNeXt-Tiny backbone with self-attention, reporting 98.73\% against 96.27\% for ConvNeXt-Tiny and 94.00\% for ResNet-50 alone---evidence that convolution and attention are complementary rather than competing, a claim this benchmark tests by evaluating both paradigms under one protocol. Parra-Medina et al.~\cite{parra2025deep} meta-analyzed deep learning for predicting oncogenic driver alterations from H\&E whole-slide images in NSCLC (pooled sensitivity 80\%, specificity 85\% for ALK), showing the field extending toward molecular inference---which raises rather than lowers the importance of a reliable subtype classifier upstream. Across this literature, however, reported accuracies remain difficult to compare, because preprocessing, augmentation, and evaluation protocols differ from study to study---precisely the confound the controlled protocol adopted here is designed to remove.

\textbf{Transformers and foundation models for pathology.}
Vision Transformers (ViTs)~\cite{dosovitskiy2020image} model the long-range spatial dependencies of whole-slide structure that convolutional receptive fields capture only indirectly, and self-supervised hierarchical ViTs learn phenotype-level representations without dense labels~\cite{ye2024cluster}. This work has since consolidated into pathology-specific foundation models trained at scale: UNI, a general-purpose encoder trained on more than 100 million tissue patches~\cite{chen2024towards}; CONCH, which aligns histopathology images with diagnostic text for zero-shot transfer~\cite{lu2024visual}; Virchow, trained for clinical-grade pan-cancer detection~\cite{vorontsov2024foundation}; and Prov-GigaPath, which extends pretraining to whole-slide context from real-world archives~\cite{xu2024whole}, with promptable segmentation models such as MedSAM~\cite{ma2024segment} a parallel development. These define the current upper bound on representational quality, but require pretraining corpora and inference budgets unavailable in most laboratory settings; this benchmark therefore characterizes what is achievable with publicly available, ImageNet-scale architectures fine-tunable on a single GPU.

\textbf{Histopathology image segmentation.}
Fully convolutional networks~\cite{long2017fcn} established the encoder--decoder paradigm underlying most biomedical segmentation models. U-Net~\cite{ronneberger2015u} became dominant through skip connections that preserve fine spatial detail; Rastogi et al.~\cite{rastogi2022gland} reported a Jaccard index of 86.4\% for gland segmentation using a U-Net-inspired network with morphological post-processing. Replacing the standard encoder with a deeper pretrained backbone (e.g., ResNet) substantially improves quality~\cite{siddique2021u, zhang2017image}, and UNet++~\cite{zhou2020unetpp} better exploits multiscale features through redesigned skip connections. DeepLabV3+~\cite{chen2018encoder}, building on the atrous framework of DeepLab~\cite{chen2017deeplab} with an atrous spatial pyramid pooling (ASPP) module, captures multi-scale context without sacrificing resolution. Object-detection architectures have likewise been repurposed for image-level classification~\cite{evaluatingYOLOevolution}, yet their systematic evaluation for histopathology remains limited.

\textbf{Integrated pipelines and remaining gaps.}
Integrated pipelines that cascade tissue classification with region segmentation for lung histopathology remain scarce~\cite{nooreldeen2021current}, although the decoupled multi-stage structure is established elsewhere in clinical imaging: in earlier work we developed an end-to-end pipeline cascading detection with downstream characterization for opportunistic vertebral fracture and Schmorl's node screening from routine CT~\cite{hasan2026end}, where stage-wise separation was what made error attribution and clinical validation tractable. Public datasets such as LC25000~\cite{borkowski2019lung} and LungHist700~\cite{Diosdado2024LungHist700} have enabled progress on isolated tasks, but benchmarks spanning a broad range of classifiers and segmentation models under a single controlled protocol, and reporting deployment cost alongside accuracy, remain limited.

\section{Methodology}
\label{sec:methodology}
\begin{figure}[!ht]
\centering
\includegraphics[width=\columnwidth]{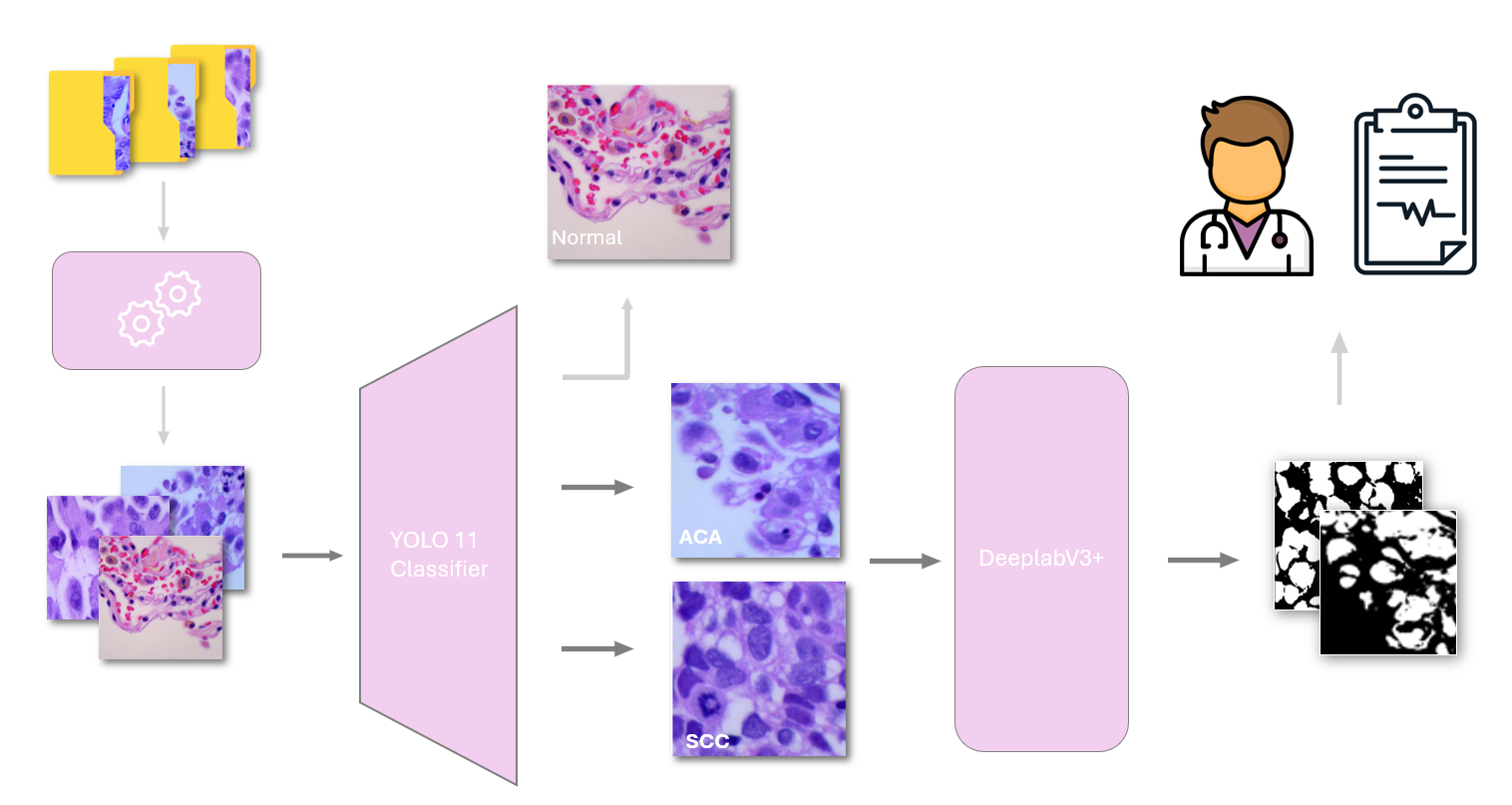}
\caption{Proposed two-stage diagnostic pipeline integrating YOLO11 for tissue classification and DeepLabV3+ for cancerous region segmentation.}
\label{fig:pipeline}
\end{figure}

This section describes the datasets, preprocessing, model architectures, and training strategies. The overall pipeline, illustrated in Fig.~\ref{fig:pipeline}, cascades a tissue classifier with a segmentation model, enabling end-to-end analysis of histopathology slides, from tissue-type prediction to spatial delineation of affected regions.

\subsection{Datasets}
\begin{figure}[!ht]
\centering
\includegraphics[width=\columnwidth]{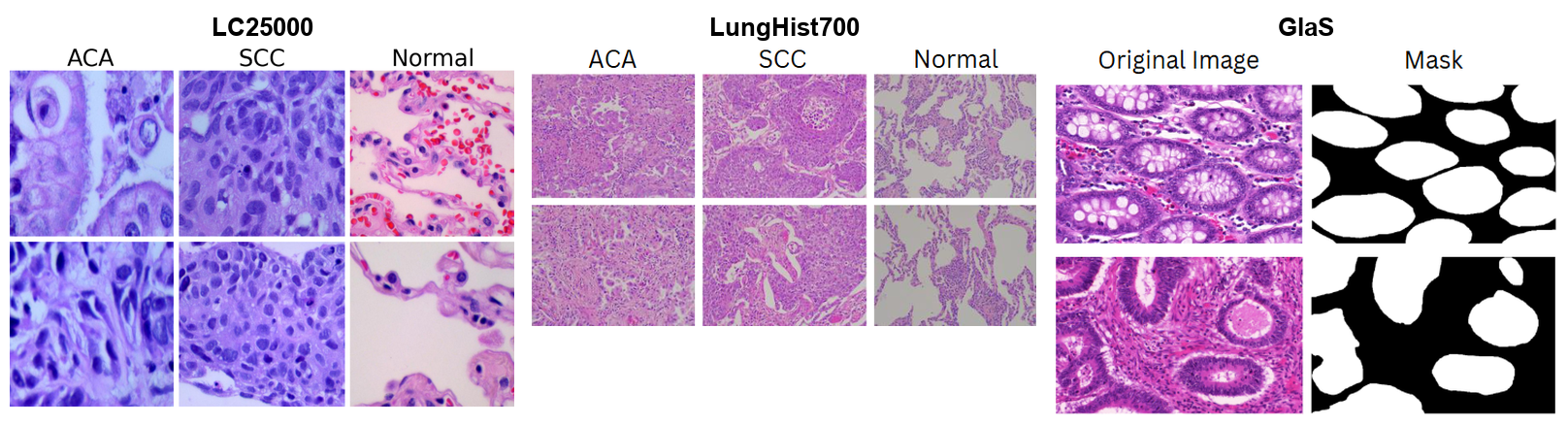}
\caption{Sample images from the three datasets. LC25000 and LungHist700 provide class-level labels (ACA, SCC, Normal); GlaS provides pixel-level segmentation masks.}
\label{fig:dataset_images}
\end{figure}

Three public histopathology datasets were used: LC25000 (25,000 images at $768\times768$), LungHist700 (700 at $1600\times1200$), and GlaS (165 at $775\times522$); the first two carry class-level labels and GlaS pixel-level masks (Fig.~\ref{fig:dataset_images}). Because LungHist700 images are captured at a lower magnification than LC25000, each of the 700 original images was subdivided into 20 non-overlapping crops, yielding 14,000 patches that approximate the field of view of the LC25000 tiles. Non-overlapping cropping mitigates exact-pixel redundancy, and because adjacent crops from the same parent image share morphological context, stratified splitting was performed at the parent-image level to prevent leakage between training and test sets. LC25000 (25,000 images) and the expanded LungHist700 (14,000 crops) were merged into a combined classification corpus of 39,000 images spanning three classes, ACA, SCC, and NOR. Merging increases both the volume and morphological diversity of the training samples, partially mitigating overfitting to the visual biases of any single source.

The GlaS Challenge dataset~\cite{sirinukunwattana2017gland}, with 165 pixel-level annotated images of colorectal glandular structures, was used to train and evaluate segmentation. Although GlaS is colorectal rather than lung-specific, it provides the pixel-level annotations required for supervised segmentation and remains the most widely adopted gland-segmentation benchmark; this cross-domain choice, necessitated by the scarcity of lung-specific pixel-level annotations, is discussed as a limitation in Section~\ref{sec:discussion}.

\subsection{Data Preprocessing}
All images were resized to $768\times768$ pixels for dimensional consistency. To reconcile staining differences between LC25000 and LungHist700, per-channel $z$-normalization (channel-wise mean subtraction and standard-deviation scaling) was applied. Data augmentation during training increased sample diversity and reduced overfitting, including random rotation ($\pm90^{\circ}$), horizontal and vertical flipping, brightness and contrast jittering ($\pm20\%$), and elastic deformation. Pixel values were subsequently normalized to $[0,1]$. The data were split into training, validation, and test sets in an 80/10/10 ratio with stratified sampling at the parent-image level to preserve class balance.

\subsection{Classification Models}
YOLO11~\cite{khanam2024yolov11, yolo11_ultralytics}, originally designed for real-time object detection, employs a deep convolutional backbone with C3k2 and C2PSA modules and spatial pyramid pooling that together aggregate features across a wide range of spatial scales. Its detection head was replaced with a fully connected classification layer, and the backbone was initialized with ImageNet-pretrained weights before fine-tuning on the merged dataset. Five additional architectures, a custom CNN, VGG16~\cite{simonyan2014very}, DenseNet~\cite{huang2017densely}, MobileNetV3~\cite{howard2019searching}, and a custom Vision Transformer, were trained under comparable conditions as baselines, spanning the convolutional, lightweight, and transformer-based paradigms. VGG16 provides a classical deep-convolutional reference with a large parameter footprint but no residual or dense connections; DenseNet introduces dense inter-layer connectivity that promotes feature reuse and alleviates vanishing gradients; and MobileNetV3 contributes a lightweight inverted-residual design with squeeze-and-excitation blocks that delivers competitive accuracy at a fraction of the parameters. For every pretrained network, the final classification layer was replaced with a three-class softmax head (ACA, SCC, NOR). The custom CNN and the custom ViT, both trained from scratch, establish non-pretrained convolutional and attention-based lower bounds respectively; the full configuration of all ten models is given in Table~\ref{tab:hyperparams}.

\subsection{Segmentation Models}
DeepLabV3+~\cite{chen2018encoder} combines an atrous-convolution encoder with an ASPP module to capture multi-scale context and a lightweight decoder that recovers fine spatial detail, making it well-suited to segmenting irregular, heterogeneous tissue regions. An ImageNet-pretrained ResNet-101 backbone was used as the encoder, with the ASPP module and decoder fine-tuned on GlaS. Three additional architectures were trained under the same conditions: U-Net~\cite{ronneberger2015u}, a ResNet-encoder U-Net~\cite{zhang2017image}, and a YOLO11 instance-segmentation variant (YOLO11-seg)~\cite{khanam2024yolov11, yolo11_ultralytics}. All segmentation models operate on $768\times768$ inputs and predict a single foreground (gland) class. U-Net serves as the canonical encoder--decoder baseline, with symmetric skip connections but a shallow, randomly initialized encoder; ResNet-U-Net retains the same decoder, while substituting a deeper ImageNet-pretrained ResNet-50 encoder, isolating the effect of encoder capacity. Unlike these encoder--decoder designs, YOLO11-seg repurposes the YOLO11 backbone with a segmentation head that produces per-instance polygon masks; the GlaS annotation masks, which encode each gland instance as a distinct label, were converted to YOLO polygon format with one contour extracted per instance, providing a markedly more parameter-efficient, detection-based alternative.

\subsection{Training Strategy}
All pretrained architectures were initialized from ImageNet weights---except YOLO11-seg, which was initialized from COCO segmentation weights---and fine-tuned using transfer learning~\cite{pan2010transfer}; Tajbakhsh et al.~\cite{tajbakhsh2016cnn} showed that fine-tuning consistently outperforms training from scratch in medical imaging. The Adam optimizer was used with cosine annealing decay and an initial learning rate of 0.001 for all models except the custom CNN, which used a lower rate of 0.0001 to stabilize training of its randomly initialized weights. All models were configured with a maximum of 100 epochs and early stopping (patience of 3 epochs on validation loss), so several pretrained models converged well before the limit. Stratified 5-fold cross-validation was performed for every model. The complete architectural and training configuration of all ten models is given in Table~\ref{tab:hyperparams}. The final pipeline integrates YOLO11 for classifying images into ACA, SCC, and NOR, followed by DeepLabV3+ for segmenting affected regions in cancerous slides. Cascading two independently optimized stages, rather than training a single multi-task network, follows established practice in clinical imaging pipelines, where decoupled stages can be validated, audited, and replaced individually as data or regulatory requirements change; we adopted the same decoupled structure for CT-based opportunistic screening~\cite{hasan2026end}, where it simplified both error attribution and clinical validation. YOLO11 and DeepLabV3+ were selected as pipeline components based on their superior and consistent performance across accuracy, F1-score, and IoU relative to all other evaluated architectures (Section~\ref{sec:experiments}).

\section{Experiments and Results}
\label{sec:experiments}

\subsection{Experimental Setup}
All experiments were run on an NVIDIA Tesla P100 GPU with 32~GB RAM, in Python using TensorFlow and PyTorch. Transfer learning was used for all architectures except the custom CNN, the custom ViT, and vanilla U-Net, which were trained from scratch. Primary per-model results correspond to the held-out test set of the 80/10/10 split, while stratified 5-fold cross-validated means and standard deviations, which estimate performance over the full corpus, are reported alongside them. Table~\ref{tab:hyperparams} summarizes the key training hyperparameters.

\begin{table*}[!t]
\caption{Complete architectural and training configuration for all ten evaluated models. Settings held constant across every model are stated here rather than repeated per row: $768\times768$ RGB input, Adam optimizer with cosine-annealing decay, a maximum of 100 epochs with early stopping at patience 3 on validation loss, and stratified 5-fold cross-validation. The initialization column also records pretraining status; WD denotes weight decay and CE cross-entropy.}
\label{tab:hyperparams}
\centering
\scriptsize
\resizebox{\textwidth}{!}{%
\begin{tabular}{|l|l|c|c|c|c|c|c|c|c|}
\hline
\textbf{Model} & \textbf{Layers / units} & \textbf{Activation} & \textbf{Initialization / pretraining} & \textbf{Dropout} & \textbf{Regularization} & \textbf{Loss} & \textbf{LR} & \textbf{Batch} & \textbf{Post-proc.} \\
\hline
\multicolumn{10}{|l|}{\textit{Stage 1 --- tissue classification}} \\
\hline
Custom CNN     & 3 conv blocks (32/64/128) + FC-256                        & ReLU          & He normal                & 0.25 (per block) & L2, $\lambda{=}10^{-3}$ & Categorical CE & $10^{-4}$ & 32 & Softmax, argmax \\
VGG16          & 13 conv + 3 FC; 3-class softmax head                      & ReLU          & ImageNet; Glorot head    & 0.5 (FC)         & WD $10^{-4}$            & Categorical CE & $10^{-3}$ & 32 & Softmax, argmax \\
DenseNet-121   & 121 layers, 4 dense blocks; 3-class head                  & ReLU          & ImageNet; Glorot head    & 0.2 (head)       & WD $10^{-4}$            & Categorical CE & $10^{-3}$ & 32 & Softmax, argmax \\
MobileNetV3    & Inverted residual + SE; 3-class head                      & ReLU/h-swish  & ImageNet; Glorot head    & 0.2 (head)       & WD $10^{-4}$            & Categorical CE & $10^{-3}$ & 32 & Softmax, argmax \\
Custom ViT     & 6 layers, 8 heads, $d{=}128$, $32^2$ patches (576 tokens) & GELU          & Trunc. normal $\sigma{=}0.02$ & 0.1         & WD $10^{-4}$            & Categorical CE & $10^{-3}$ & 16 & Softmax, argmax \\
YOLO11 (s)     & C3k2 + C2PSA + SPPF; FC classifier head                   & SiLU          & ImageNet; He head        & 0.0              & WD $5{\times}10^{-4}$   & Categorical CE & $10^{-3}$ & 32 & Softmax, argmax \\
\hline
\multicolumn{10}{|l|}{\textit{Stage 2 --- region segmentation}} \\
\hline
U-Net          & 4 down/up stages, 64--1024 ch., skip connections          & ReLU          & He normal                & 0.5 (bottleneck) & WD $10^{-4}$            & BCE + Dice     & $10^{-3}$ & 16 & Sigmoid @ 0.5 \\
ResNet-U-Net   & ResNet-50 encoder + U-Net decoder                         & ReLU          & ImageNet enc.; He dec.   & 0.5 (bottleneck) & WD $10^{-4}$            & BCE + Dice     & $10^{-3}$ & 16 & Sigmoid @ 0.5 \\
DeepLabV3+     & ResNet-101 + ASPP + decoder (output stride 16)            & ReLU          & ImageNet enc.; He dec.   & 0.1 (ASPP)       & WD $10^{-4}$            & BCE + Dice     & $10^{-3}$ & 16 & Sigmoid @ 0.5 \\
YOLO11-seg (n) & YOLO11 backbone + polygon mask head                       & SiLU          & COCO; He head            & 0.0              & WD $5{\times}10^{-4}$   & BCE + DFL + CIoU & $10^{-3}$ & 8 & Polygon $\rightarrow$ raster; NMS IoU 0.7 \\
\hline
\end{tabular}}
\end{table*}

\subsection{Classification Results}
\begin{figure}[!ht]
\centering
\includegraphics[width=\columnwidth]{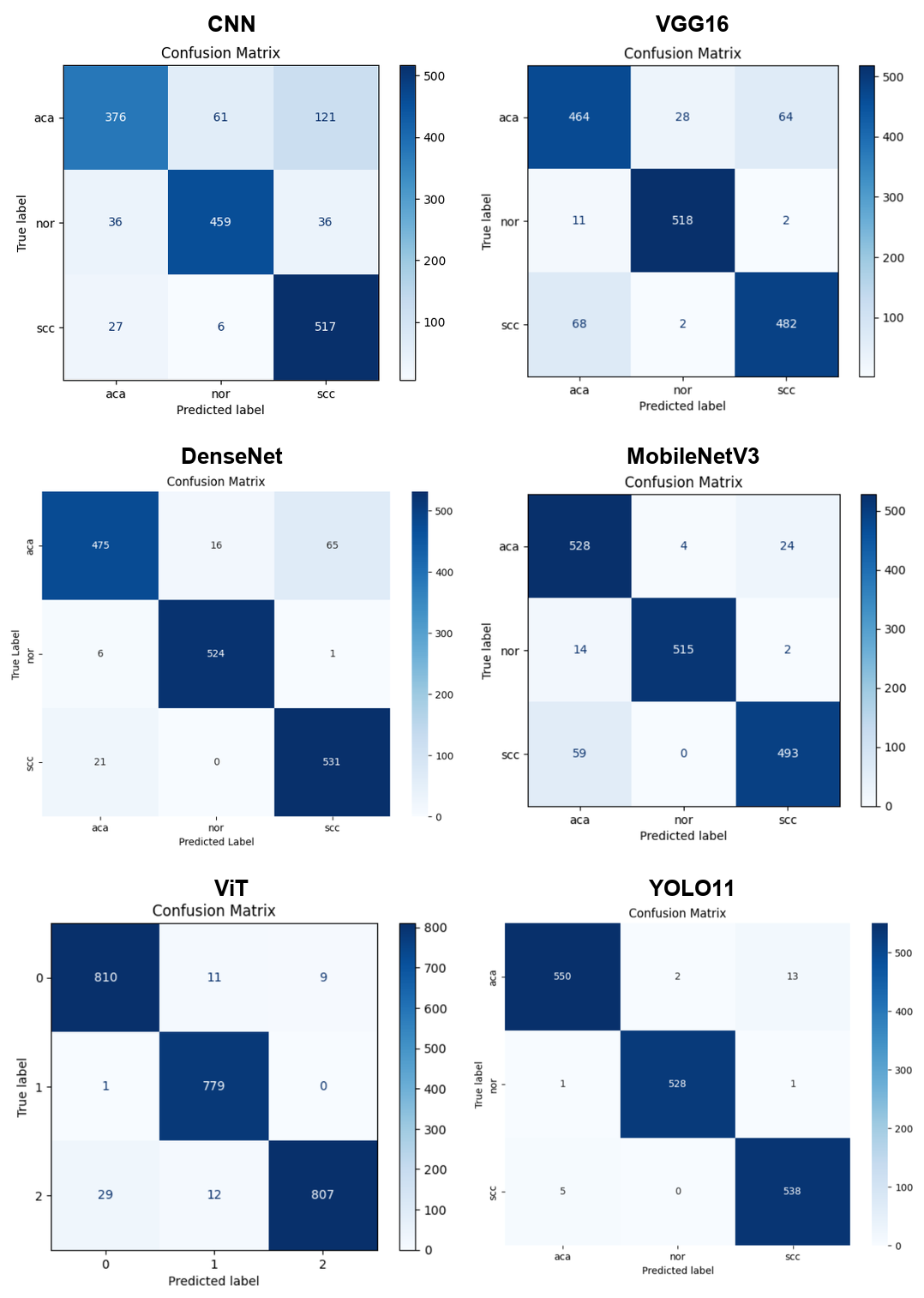}
\caption{Confusion matrices on the merged test set for all six classifiers. Classes appear as aca/nor/scc, or as 0/1/2 in the same order for the ViT panel. YOLO11 and ViT show the fewest off-diagonal errors, with the ACA vs.\ SCC distinction being the hardest across models.}
\label{fig:confusion}
\end{figure}

\begin{figure}[!ht]
\centering
\includegraphics[width=\columnwidth]{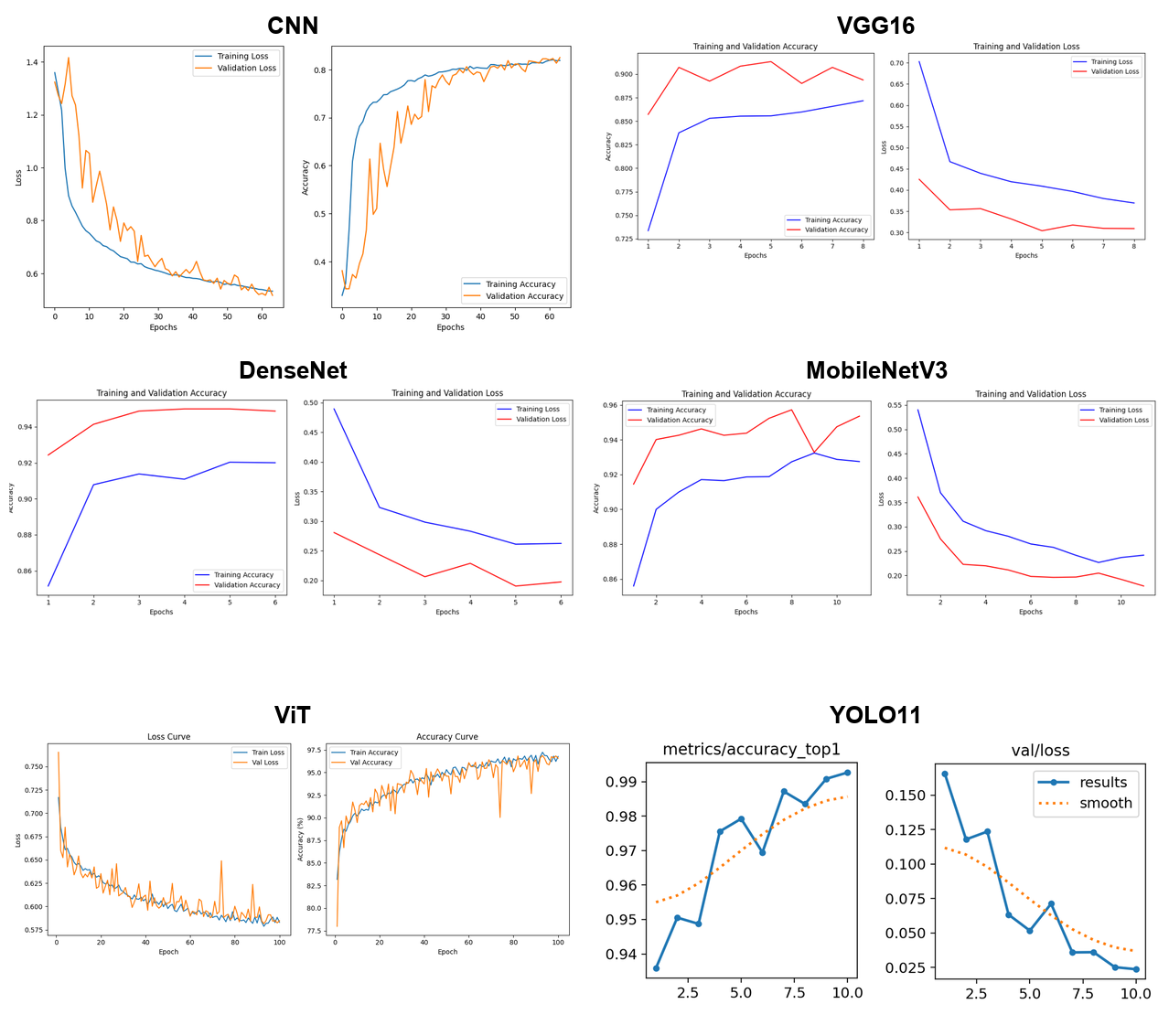}
\caption{Training and validation loss/accuracy curves for the six classifiers. Pretrained models converge within a few epochs; YOLO11 converges in approximately 10 epochs.}
\label{fig:curves}
\end{figure}

Table~\ref{tab:classification_results} reports single-split accuracy and macro F1-score alongside 5-fold cross-validation means for the six classifiers; confusion matrices and training curves appear in Figs.~\ref{fig:confusion} and~\ref{fig:curves}. A clear hierarchy emerges: architectures with deeper pretrained backbones and multi-scale feature aggregation consistently outperform shallower or non-pretrained models. The custom CNN, lacking pretrained features, underfits and caps at 83.5\%, whereas YOLO11 achieves the best overall performance (98.38\%, F1\,=\,0.98), converging in approximately 10 epochs owing to its effective pretrained backbone, followed closely by the custom ViT (97.1\%). This supports the hypothesis that multi-scale convolutional backbones designed for detection transfer effectively to histopathology classification when the detection head is replaced by a classification layer. The cross-validated means are closely aligned with the single-split results, and YOLO11 exhibits the smallest standard deviation ($\pm0.35\%$), indicating robust performance across data partitions.

Examining the models individually, the from-scratch CNN converges with closely matched training, validation, and test accuracies, so underfitting rather than variance is its bottleneck. VGG16 trails the denser networks, consistent with the absence of residual or dense connections, while DenseNet and MobileNetV3 perform almost identically---the latter at much lower computational cost. The custom ViT's 97.1\% despite training from scratch is evidence that self-attention captures long-range spatial dependencies of tissue architecture even at moderate data scale. YOLO11's confusion matrix (Fig.~\ref{fig:confusion}) confines almost all residual error to the ACA--SCC off-diagonal while separating normal tissue almost perfectly, a pattern shared to a lesser degree by every model.

\begin{table}[tb]
\caption{Classification performance on the merged LC25000--LungHist700 test set, with 5-fold cross-validation (mean $\pm$ std).}
\label{tab:classification_results}
\centering
\resizebox{\columnwidth}{!}{
\begin{tabular}{|l|c|c|c|c|}
\hline
\textbf{Model} & \textbf{Acc. (\%)} & \textbf{Macro F1} & \textbf{CV Acc. (\%)} & \textbf{CV F1} \\
\hline
CNN & 83.50 & 0.84 & $83.12\pm1.24$ & $0.83\pm0.01$ \\
VGG16 & 88.40 & 0.89 & $88.05\pm0.87$ & $0.88\pm0.01$ \\
DenseNet & 92.56 & 0.93 & $92.31\pm0.68$ & $0.92\pm0.01$ \\
MobileNetV3 & 92.80 & 0.93 & $92.54\pm0.73$ & $0.93\pm0.01$ \\
ViT & 97.10 & 0.97 & $96.88\pm0.51$ & $0.97\pm0.01$ \\
YOLO11 & \textbf{98.38} & \textbf{0.98} & $\mathbf{98.21\pm0.35}$ & $\mathbf{0.98\pm0.00}$ \\
\hline
\end{tabular}
}
\end{table}

\subsubsection{Per-Class Analysis}
The per-class ordering is identical across all six models: normal tissue (NOR) is the most readily distinguished class (F1 from 0.90 for the custom CNN to 0.99 for ViT and YOLO11), while the clinically critical ACA vs.\ SCC distinction is consistently the hardest, reflecting morphological overlap between the two subtypes. SCC is the weakest class for every model, ranging from F1\,=\,0.80 (CNN) through 0.85 (VGG16), 0.90 (DenseNet), 0.91 (MobileNetV3), and 0.96 (ViT) to 0.97 (YOLO11). YOLO11 maintains balanced performance across all three classes (ACA 0.98, SCC 0.97, NOR 0.99), and its lowest per-class F1 still matches or exceeds the macro F1 of every other evaluated model. This balance is clinically significant, as misclassifying one cancer subtype as another can lead to inappropriate treatment.

\subsection{Segmentation Results}
\begin{figure}[!ht]
\centering
\includegraphics[width=\columnwidth]{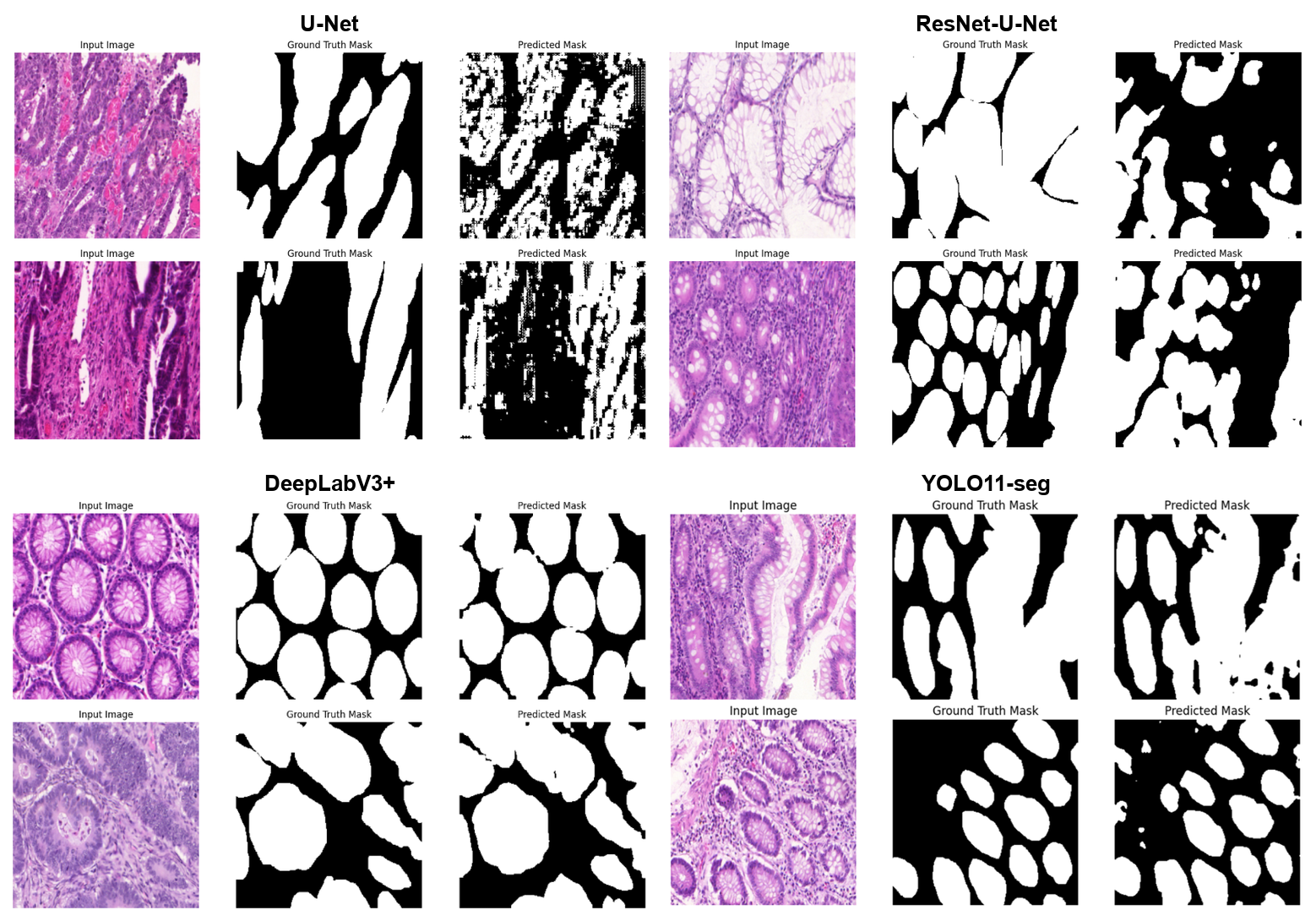}
\caption{Qualitative segmentation outputs on GlaS, two test cases per model, each shown as input image, ground-truth mask, and predicted mask. U-Net's predictions are visibly fragmented, whereas DeepLabV3+ and YOLO11-seg recover coherent gland boundaries at very different parameter budgets.}
\label{fig:segmentation}
\end{figure}

Table~\ref{tab:segmentation_results} reports IoU and Dice for the four segmentation architectures on the GlaS test set, with sample outputs in Fig.~\ref{fig:segmentation}. The pretrained encoder backbone is the dominant factor: ResNet-U-Net improves IoU by 0.27 over vanilla U-Net, whose limited encoder capacity on a small dataset without pretraining yields the weakest result. DeepLabV3+ achieves the best score (IoU\,=\,0.80, Dice\,=\,0.89) through its ASPP-based multi-scale aggregation, with the lowest cross-validation variance ($\pm0.02$ IoU). Notably, YOLO11-seg attains a competitive IoU of 0.79 (Dice\,=\,0.88)---second only to DeepLabV3+ and above ResNet-U-Net---with only 2.9M parameters, roughly $14\times$ fewer than DeepLabV3+.

The qualitative outputs in Fig.~\ref{fig:segmentation} are consistent with the quantitative ranking. Vanilla U-Net produces fragmented masks with jagged, discontinuous boundaries, reflecting the limited capacity of its non-pretrained encoder trained on only 165 images; substituting a pretrained ResNet encoder restores coherent glandular contours, which accounts for the 0.27 IoU gain. DeepLabV3+ yields the cleanest boundaries, its ASPP module resolving glands across a wide range of sizes without loss of spatial resolution, whereas YOLO11-seg, predicting per-instance polygon masks rather than a dense pixel map, occasionally merges adjacent glands but rarely misses them. The low cross-validation variance indicates this ranking is stable rather than an artifact of a particular split.

\begin{table}[tb]
\caption{Segmentation performance on the GlaS test set, with 5-fold cross-validation (mean $\pm$ std).}
\label{tab:segmentation_results}
\centering
\resizebox{\columnwidth}{!}{
\begin{tabular}{|l|c|c|c|c|}
\hline
\textbf{Model} & \textbf{IoU} & \textbf{Dice} & \textbf{CV IoU} & \textbf{CV Dice} \\
\hline
U-Net~\cite{ronneberger2015u} & 0.51 & 0.68 & $0.50\pm0.04$ & $0.67\pm0.04$ \\
ResNet-U-Net~\cite{zhang2017image} & 0.78 & 0.88 & $0.77\pm0.03$ & $0.87\pm0.02$ \\
YOLO11-seg & 0.79 & 0.88 & $0.78\pm0.03$ & $0.87\pm0.03$ \\
DeepLabV3+~\cite{chen2018encoder} & \textbf{0.80} & \textbf{0.89} & $\mathbf{0.79\pm0.02}$ & $\mathbf{0.88\pm0.02}$ \\
\hline
\end{tabular}
}
\end{table}

\subsection{Computational Cost}
Table~\ref{tab:compute} reports parameters, GFLOPs, and single-image inference latency at $768\times768$ resolution. VGG16 is by far the most parameter-heavy model (134.3M) yet delivers only moderate accuracy, while MobileNetV3 reaches higher accuracy with $25\times$ fewer parameters. YOLO11 offers the best accuracy--efficiency trade-off among classifiers (9.4M parameters, 13.5 GFLOPs, 6\,ms latency). For segmentation, DeepLabV3+ incurs the highest cost but yields the best IoU, whereas YOLO11-seg attains nearly equivalent IoU with only 2.9M parameters, making it attractive for latency-sensitive deployment.

\begin{table}[tb]
\caption{Computational cost of all models. Latency is single-image inference at $768\times768$ on an NVIDIA Tesla P100 GPU.}
\label{tab:compute}
\centering
\resizebox{0.9\columnwidth}{!}{
\begin{tabular}{|l|r|r|r|}
\hline
\textbf{Model} & \textbf{Params (M)} & \textbf{GFLOPs} & \textbf{Latency (ms)} \\
\hline
CNN & 0.5 & 1.2 & 4 \\
VGG16 & 134.3 & 96.4 & 18 \\
DenseNet-121 & 7.0 & 17.2 & 12 \\
MobileNetV3 & 5.4 & 1.8 & 5 \\
ViT & 10.8 & 19.6 & 14 \\
YOLO11 & 9.4 & 13.5 & 6 \\
\hline
U-Net & 31.0 & 54.8 & 26 \\
ResNet-U-Net & 32.5 & 68.2 & 33 \\
DeepLabV3+ & 40.4 & 84.6 & 38 \\
YOLO11-seg & 2.9 & 10.4 & 8 \\
\hline
\end{tabular}
}
\end{table}

\subsection{Ablation Studies}
Three ablations quantify key design decisions. \textit{Dataset merging:} training YOLO11 on the merged corpus yields 98.38\% accuracy, a 0.56 percentage-point improvement over LC25000 alone (97.82\%) and a 6.95-point improvement over LungHist700 alone (91.43\%), confirming that merging increases sample diversity and volume. \textit{Data augmentation:} disabling augmentation reduces YOLO11 accuracy from 98.38\% to 96.91\% (a 1.47-point drop), demonstrating its contribution to robustness. \textit{Encoder backbone:} for DeepLabV3+ on GlaS, a MobileNetV2 encoder yields IoU\,=\,0.74, ResNet-50 yields 0.78, and ResNet-101 yields the best 0.80 (Dice\,=\,0.89), indicating diminishing returns from deeper encoders on this small dataset.

\subsection{Comparison with Prior Work}
Published results on these benchmarks are not directly comparable to ours, and the reasons are instructive. On lung histopathology, Abbas et al.~\cite{abbas2020histopathological} report F1-scores of 97.3--99.9\% across six pretrained CNNs and Humayun et al.~\cite{humayun2022transfer} 98.83\% with VGG16, both on LC25000 alone and each tuned to that single source; Ochoa-Ornelas et al.~\cite{ochoa2025robust} reach 99.39\% on its full five-class version. Our 98.38\% is obtained on a three-class corpus merging LC25000 with lower-magnification LungHist700 crops, under a protocol held fixed across ten architectures rather than optimized per model. On GlaS, Rastogi et al.~\cite{rastogi2022gland} report a Jaccard index of 86.4\% (Dice 92.4\%) using a U-Net-inspired network with morphological post-processing, above our DeepLabV3+ (IoU\,=\,0.80, Dice\,=\,0.89), which applies no task-specific post-processing. The aim here is therefore not the highest figure on either benchmark: single-source accuracies obtained under per-model tuning are not commensurable with results from a uniform protocol, which is what motivates the controlled comparison reported above and what makes its conclusions relative---which paradigm transfers, where pretraining matters, and what accuracy costs.

\section{Discussion}
\label{sec:discussion}
The results reveal a consistent picture across both stages. The $\sim$15 percentage-point gap between the non-pretrained custom CNN and YOLO11 underscores the decisive role of transfer learning in a domain pairing high visual complexity with limited annotated data, and the progression through VGG16, DenseNet, and MobileNetV3 shows that architectural advances compound that benefit. YOLO11's leading accuracy despite its object-detection origins, together with the competitive YOLO11-seg, indicates that detection-oriented backbones transfer effectively to computational pathology and offer the most favorable accuracy--efficiency profiles for latency-sensitive deployment. On segmentation, the 0.27 IoU jump from vanilla U-Net to ResNet-U-Net confirms that pretrained encoder capacity---rather than decoder sophistication---dominates on small datasets, with the backbone ablation showing diminishing returns beyond ResNet-50, so data volume rather than model capacity is the binding constraint. The principal limitations are that the segmentation stage is trained on GlaS rather than scarce lung-specific pixel annotations---so clinical use would require retraining on lung data---and that robustness to inter-laboratory staining variation and formal significance testing remain to be established; nonetheless, the cross-validated, ablated, and cost-profiled results support the pipeline as a competitive, well-characterized baseline.

The design is not specific to lung histopathology. Because the two stages are decoupled and share only a common preprocessing and evaluation protocol, the same cascade transfers to any domain in which a categorical decision precedes spatial delineation: we have instantiated an equivalent structure on routine CT for opportunistic vertebral fracture and Schmorl's node screening~\cite{hasan2026end}. That the structure holds across modalities as different as whole-slide microscopy and volumetric CT suggests the transferable element is the selection methodology rather than dataset-specific tuning; the absolute figures should not be assumed to carry over, and another modality would require re-running the selection procedure on target-domain data.

Two considerations, previewed in Section~\ref{sec:introduction}, condition any clinical use of these results. First, we have shown that deep classifiers remain vulnerable to adaptive attacks conventional metrics do not surface, and proposed reinforcement-learning-based hardening as a defense~\cite{hasan2026defending}---an axis orthogonal to the accuracy, IoU, and latency reported here. Second, a prediction becomes clinically actionable only within a system that exposes its uncertainty and provenance to the clinician; the trustworthiness requirements we identify for large-language-model agents in healthcare~\cite{hasan2026toward}---verifiable grounding, calibrated abstention, and auditable decision traces---apply equally to an image-based diagnostic pipeline.

\section{Conclusion}
This work presents a systematic multi-architecture benchmark and an integrated two-stage pipeline for automated classification and segmentation of lung cancer histopathology. Its contribution is comparative evidence gathered under a single controlled protocol rather than a new architecture. Among six classifiers, YOLO11 achieves the best test accuracy of 98.38\% (5-fold CV: $98.21\pm0.35\%$) on the merged 39,000-image LC25000--LungHist700 dataset, with balanced per-class performance on the clinically critical ACA vs.\ SCC distinction. Among four segmentation architectures on GlaS, DeepLabV3+ attains the best IoU of 0.80, while YOLO11-seg is competitive (0.79) at $14\times$ fewer parameters. Ablations confirm the value of dataset merging, augmentation, and encoder backbone selection. Future work will retrain the segmentation stage on lung-specific pixel-level annotations, incorporate stain normalization and domain adaptation for cross-laboratory robustness, and extend the benchmark with formal statistical comparison and pathology foundation-model backbones. A further direction is to pair the pipeline's predictions with a knowledge-augmented retrieval layer over curated pathology references, letting a clinician interrogate the supporting literature alongside the model's output; our retrieval-grounded generation system~\cite{hasan2025kag} offers a basis for that interface.

\bibliographystyle{IEEEtran}
\bibliography{references}

\end{document}